\documentclass[11pt,a4paper]{article}
\usepackage[hyperref]{eacl2021}
\usepackage{times}
\usepackage{latexsym}

\usepackage{microtype}
\usepackage{hyperref}
\usepackage{booktabs}
\usepackage{graphicx}
\usepackage{caption}
\usepackage{subcaption}
\usepackage{enumitem}

\usepackage{arabtex}
\usepackage{utf8}
\usepackage{latexsym}
\usepackage{textcomp}

\aclfinalcopy 
\def\aclpaperid{31} 

\renewenvironment{abstract}%
		 {\centerline{\large\bf Abstract}%
		  \begin{list}{}%
		     {\setlength{\rightmargin}{0.6cm}%
		      \setlength{\leftmargin}{0.6cm}}%
		   \item[]\ignorespaces
		   \smalll
		   }%
		 {\unskip\end{list}}

\title{\textsc{AraGPT2}: Pre-Trained Transformer \\for Arabic Language Generation}

\author{Wissam Antoun \and Fady Baly \and Hazem Hajj \\
        American University of Beirut\\
        \{wfa07, fbg06, hh63\}@aub.edu.lb}

\date{}

\begin{document}
\maketitle
\begin{abstract}
Recently, pre-trained transformer-based architectures have proven to be very efficient at language modeling and understanding, given that they are trained on a large enough corpus.
Applications in language generation for Arabic are still lagging in comparison to other NLP advances primarily due to the lack of advanced Arabic language generation models.
In this paper, we develop the first advanced Arabic language generation model, AraGPT2, trained from scratch on a large Arabic corpus of internet text and news articles.
Our largest model, \textsc{AraGPT2-mega}, has 1.46 billion parameters, which makes it the largest Arabic language model available.
The \textsc{mega} model was evaluated and showed success on different tasks including synthetic news generation, and zero-shot question answering.
For text generation, our best model achieves a perplexity of 29.8 on held-out Wikipedia articles.
A study conducted with human evaluators showed the significant success of AraGPT2-mega in generating news articles that are difficult to distinguish from articles written by humans.
We thus develop and release an automatic discriminator model with a 98\% percent accuracy in detecting model-generated text.
The models are also publicly available\footnote{Pretrained variants of \textsc{AraGPT2} (base, medium, large, mega) and discriminator are publicly available on \href{https://github.com/aub-mind/arabert/tree/master/aragpt2}{github.com/aub-mind/arabert/tree/master/aragpt2}}, hoping to encourage new research directions and applications for Arabic NLP.
\end{abstract}

\section{Introduction}
Few years ago, Natural language processing (NLP) was revolutionized with the introduction of multi-head self-attention transformer architecture~\cite{vaswani2017attention}.
The transformer achieved superior performance compared to recurrent neural networks several NLP tasks including machine translation, sentence classification with BERT~\cite{devlin2019bert}, and ELECTRA~\cite{clark2020electra}, and sentence completion with GPT-2~\cite{radford2019language}, GROVER~\cite{zellers2019defending}, and CTRL~\cite{keskar2019ctrl}.
Recent works have shown that larger models pre-trained on larger datasets can further improve performance i.e. RoBERTa~\cite{liu2019roberta}, and XLM-R~\cite{conneau2019unsupervised}.

On the other hand, work on Arabic language modeling has mostly targeted natural language understanding (NLU) by pre-training transformer-based models using the Masked Language Modeling (MLM) task i.e. \textsc{AraBERT}~\cite{antoun2020arabert}.
In contrast, Arabic text generation or causal language modeling hasn't received much attention. 
Few works such as hULMonA~\cite{eljundi2019hulmona} used next word prediction as a pre-training task in for transfer learning in Arabic text classification.
\cite{khooligpt2} and \cite{doiron_2020} leveraged the existing GPT2 English model and adapted it for Arabic using text from the Arabic Wikipedia dumps, which is sub-optimal for Arabic.

In this paper, the first advanced language generation models built from the grounds up on Arabic language have been developed.  
The process of pre-training \textsc{AraGPT2}, a GPT-2 transformer model for the Arabic language is described.
The model comes in 4 size variants: \textbf{base} (135M\footnote{Million Parameters}), \textbf{medium} (370M), \textbf{large} (792M) and \textbf{mega} (1.46B\footnote{Billion Parameters}), which allows the exploration of \textsc{AraGPT2} in multiple applications with different data availability and computational constraints. 
The perplexity measure is used to automatically evaluate \textsc{AraGPT2}.
Furthermore,  a human-based evaluation is provided, which highlights the ability of \textsc{AraGPT2} to deceive human evaluators.
Finally, an \textsc{AraELECTRA}~\cite{antoun2020araelectra} based detector is developed and released. It is able to consistently identify news articles written by \textsc{AraGPT2}.
Making such powerful models publicly available to the Arabic research community enables research in rising Arabic NLP fields i.e Conversational Agents~\cite{naous-etal-2020-empathy}, Detection of Automatic News Generation Detection~\cite{harrag-etal-2020-bert}...

Our contributions can be summarized as follows:
\begin{itemize}[noitemsep,topsep=0pt]
    \item A methodology to pre-train a billion-size GPT2 model on a large-scale Arabic corpus.
    \item An automatic discriminator that achieves a 98\% accuracy in detecting model-generated synthetic text.
    \item The four variants of \textsc{AraGPT2} are released on popular NLP libraries, along with the automatic \textsc{AraGPT2} discriminator.
\end{itemize}

The rest of the paper is structured as follows. Section~\ref{sec:rel} provides a concise review of previous literature on Arabic language modeling.
Section~\ref{sec:method} details the methodology used in developing \textsc{AraGPT2}.
Section~\ref{sec:eval} describes the experimental setup, evaluation procedures and results.
In addition, the approach to build a machine-generated text discriminator is presented in Section~\ref{sec:disc}.
Finally, a conclusion of the work and implications are mentioned in Section~\ref{sec:conc}.
\section{Related Works}
\label{sec:rel}
\subsection{English and Non-Arabic Language modeling}
GPT-1~\cite{radford2018improving} showed that Causal Language Modeling\footnote{This is the regular Language Modeling objective where the model learns the probability of a word given the previous context. The CLM acronym is used to distinguish from masked language modeling (MLM).} is an effective pre-training technique that improves a model's generalization capabilities.
GPT-2 then showed that using a larger model trained on a larger dataset surpasses the state-of-the-art of many tasks in a zero-shot setting, where a model solves a task without receiving any training on that task. Taking the scaling approach to the extreme led to the creation of GPT-3~\cite{brown2020language}, with 175 billion parameter model, also trained with CLM using terabytes of internet text.
GPT-3 explored the idea of few-shot learning, where a model is given examples from a new task as a text prompt, which unlocks new capabilities at test time.
It was later shown that a carefully designed GPT-3 prompt allows the model to generate website designs, scramble/unscramble words...
The advantage of scaling model sizes and training datasets comes with drawbacks, particularly the high computational cost, in addition to the huge corpora required for pre-training.
It was estimated that training GPT-2 and GPT-3 costs \$43K and \$4.6M respectively, without any hyper-parameter tuning.
These drawbacks restricted the availability of large pre-trained models to English mainly and a handful of other languages i.e. ruGPT3\footnote{https://github.com/sberbank-ai/ru-gpts/} for Russian, and Chinese 1.5B GPT2~\cite{GPT2-ML}.
\subsection{Arabic Language modeling}

Work on Arabic causal language modeling has been mostly limited to automatic speech recognition (ASR) systems.
Since the language modeling component in ASR systems is a key module that ensures that the output text adheres with the statistical structure of language.
Work on Arabic language models in ASR systems has mostly relied on N-grams language models.
\cite{ali2014complete} built an N-grams language model (LM) using GALE training data transcripts of 1.4M words.
More recent work in Arabic ASR implemented a recurrent neural network as an LM, using 130M tokens, and achieved a perplexity of 481 compared to 436 for a 4-gram LM~\cite{khurana2019darts}.
\citet{hamed2017building} developed a code-switched Arabic-English language model using tri-gram LM and provided performance superior compared to two separate monolingual LMs. 
The code-switched LM was trained on 2.3M sentences or 13M words and achieved a perplexity of 275.

With the rising popularity of transfer learning in NLP, Arabic CLM was used as a pre-training task for an Arabic universal LM, hULMonA~\cite{eljundi2019hulmona}.
The model was then fine-tuned on different downstream text classification tasks.
hULMonA is a 3 stack of AWD-LSTM\footnote{ASGD Weight-Dropped LSTM} layers~\cite{howard2018universal}, trained on 600K Wikipedia article pre-segmented using the MADAMIRA Arabic morphological analyzer and disambiguator~\cite{pasha-etal-2014-madamira}.  

Masked Language Modeling (MLM) has been useful as a pre-training task for several Arabic NLU models.
Masked Language Modeling (MLM) is a slightly different objective than CLM that requires a system to predict a masked word within a sequence compared to CLM which predicts the missing word at the end of a sequence.
MLM was used in models such as \textsc{AraBERT}~\cite{antoun2020arabert}, Arabic-BERT~\cite{safaya-etal-2020-kuisail}, Arabic-ALBERT\footnote{https://github.com/KUIS-AI-Lab/Arabic-ALBERT/}, GigaBERT~\cite{lan2020gigabert}, MarBERT~\cite{abdul2020toward}, and QARiB~\cite{chowdhury-etal-2020-improving-arabic}.
Only two works have attempted to create an Arabic transformer causal language model.
\citet{khooligpt2} and \citet{doiron_2020} finetuned the OpenAI GPT2-base model on Arabic Wikipedia, which was mainly trained on English text.
\citet{doiron_2020} also continued training on a collection of dialectal Arabic datasets, in order to create a dialectal Arabic GPT2. 
While this approach has shown the capability to generate Arabic text, it is sub-optimal for Arabic and is useful in cases where the training data is scarce.

Our proposed model is hence, the first Arabic transformer-based causal language model trained from scratch on the largest Arabic corpora available at the time of writing.
\section{\textsc{AraGPT2}: Methodology}
\label{sec:method}
\begin{table*}[!t]
\centering
\resizebox{0.85\textwidth}{!}{
\begin{tabular}{l|ccccccc}
\hline
\multicolumn{1}{c|}{\textbf{Model}} & \textbf{Size} & \textbf{Architecture} & \textbf{Context Size} & \textbf{Emb. Size} & \textbf{Heads} & \textbf{Layers} & \textbf{Optimizer} \\ \hline
\textbf{Base} & 135M & GPT2 & 1024 & 768 & 12 & 12 & LAMB \\
\textbf{Medium} & 370M & GPT2 & 1024 & 1024 & 16 & 24 & LAMB \\
\textbf{Large} & 792M & GROVER & 1024 & 1280 & 20 & 36 & Adafactor \\
\textbf{Mega} & 1.46B & GROVER & 1024 & 1536 & 24 & 48 & Adafactor \\ \hline
\end{tabular}
}
\caption{\textsc{AraGPT2} model variants with sizes, architecture and optimizer\label{GPT2variant}}
\end{table*}

\begin{table*}[t]
\centering
\resizebox{0.65\textwidth}{!}{
\begin{tabular}{l|ccccl}
\hline
\multicolumn{1}{c|}{\textbf{Model}} & \textbf{Batch Size} & \textbf{Learning Rate} & \textbf{Steps} & \textbf{Time (days)} & \textbf{PPL} \\ \hline
\textbf{Base} & 1792 & 1.27e-3 & 120K & 1.5 & 55.8 \\
\textbf{Medium*} & 80 & 3e-4 & 1M & 23 & 45.7 \\
\textbf{Large} & 256 & 1e-4 & 220K & 3 & 36.6 \\
\textbf{Mega} & 256 & 1e-4 & 780K & 9 & \textbf{29.8}\\ \hline
\end{tabular}
}
\caption{\textsc{AraGPT2} training details and validation perplexity.
*\textbf{Medium} was trained on a TPUv3-8 with a small batch size, since the model was not converging with a large batch size\label{gpt2res}}
\end{table*}

\textsc{AraGPT2} is a stacked transformer-decoder model trained using the causal language modeling objective.
The model is trained on 77GB of Arabic text.
\textsc{AraGPT2} comes in four variants as detailed in Table~\ref{GPT2variant}, with the smallest model, \textbf{base}, having the same size as \textsc{AraBERT}-base which makes it accessible for the larger part of researchers.
Larger model variants (\textbf{medium}, \textbf{large}, \textbf{xlarge}) offer improved performance but are harder to fine-tune and computationally more expensive.
The \textsc{AraGPT2}-detector is based on the pre-trained \textsc{AraELECTRA} model fine-tuned on the synthetically generated dataset.
More details on the training procedure and dataset are provided in the following sections.

\subsection{Model}
\textsc{AraGPT2} closely follows GPT2's variant architectures and training procedure.
Table~\ref{GPT2variant} shows the different model sizes, number of heads, number of layers, parameter count, and optimizer used for each model variant.
All models are trained with context sizes of 1024 tokens.
The LAMB~\cite{you2019large} optimizer is used in the \textbf{base} and \textbf{medium} models only, since it allows using large batch sizes without worrying about training divergence.
Using LAMB and Adam~\cite{kingma2014adam} to train the \textbf{large} and \textbf{mega} variants isn't possible on TPUv3 due to the optimizer's high memory requirements, since memory cost scales linearly with the number of parameters.
The limitations were overcome by following the training procedure of the GROVER model~\cite{zellers2019defending} by using the Adafactor optimizer~\cite{shazeer2018adafactor}, which reduces memory requirements by factoring the second-order momentum parameters into a tensor product of two vectors.
The GROVER architecture was also used instead of GPT2's, in which the layer normalization order in the transformer block is changed.

\subsection{Dataset}
The training dataset is a collection of the publicly available Arabic corpora listed below:
\begin{itemize}[noitemsep,topsep=0pt]
    \item The unshuffled OSCAR corpus~\cite{ortiz-suarez-etal-2020-monolingual}.
    \item The Arabic Wikipedia dump from September 2020.
    \item The 1.5B words Arabic Corpus~\cite{el20161}.
    \item The OSIAN corpus~\cite{zeroual2019osian}.
    \item News articles provided by As-safir newspaper.
\end{itemize}

\paragraph{Preprocessing}
First, the corpus was filtered by removing short documents with less than 3 sentences, and documents with more than 20\% repeated sentences.
URLs, emails, and user mentions were also replaced with special tokens.
All diacritics, and elongations were removed as well, while punctuation and non-alphabetic characters were padded with white-spaces.
Moreover, the \texttt{`\textless |endoftext|\textgreater'} token is appended at the end of each document.
The total dataset size is 77GB with 8.8B words\footnote{Word count was done after preprocessing, where white space is inserted before and after punctuations, brackets, numbers... which increased the total word count}.
The majority of the training data is comprised of Arabic news article, which is mostly written in MSA.
The corpus also contains a small set of English words i.e. named entities, which are kept without lower-casing.
Subsequently, a Byte-level byte-pair-encoding (BPE) tokenizer is trained with 64000 vocabulary size on all of our preprocessed dataset, using the optimized BPE implementation from the HuggingFace library~\cite{wolf-etal-2020-transformers}.
Finally, the BPE encoding is applied on the preprocessed dataset, which results in a total of 9.7M training examples with 1024 sub-word tokens each.

\section{Experiments and Evaluation}
\label{sec:eval}
\subsection{Pre-training Setup}
All models were trained on a TPUv3-128 slice\footnote{TPUv3-128 has a total of 2TB of HBM memory with 16GB per core. TPUs were freely provided by the TFRC program.} with different batch sizes and the total number of steps as shown in Table~\ref{gpt2res}.
\textbf{Base} and \textbf{mega} were trained for approximately 20 epochs, while \textbf{medium} and \textbf{large} were trained for 10 and 6 epochs respectively, due to TPU access limitations.

\subsection{Numerical Evaluation}
For the validation dataset, the Arabic Wikipedia articles that were published after August 2020 were used, since older articles were included in the September Wikipedia dump.
The perplexity score was selected as a numerical evaluation metric since it measures the degree of 'uncertainty' a model has assigning probabilities to the test text.
Table~\ref{gpt2res} shows that, unsurprisingly, validation perplexity keeps improving with larger model sizes.
In fact, the model is still under-fitting the validation set from Wikipedia.
The generation capabilities of the different variants of \textsc{AraGPT2} is illustrated through the selected examples in Appendix~\ref{appendixa}.

\subsection{Zero-Shot Evaluation}
During zero-shot task evaluation, the model is only given a natural language instruction to motivate and ground the task, without any back-propagation happening.
The task of searching and finding the best input prompt, also known as ``prompt engineering'', is hard.
Since the search space is practically infinite, and the performance is highly sensitive to changes in the prompt.
The zero-shot performance of \textsc{AraGPT2}-Mega is evaluated on two tasks, question answering, and translation.
\textsc{AraGPT2-mega} correctly answers 25\% of the trivia questions but fails in English-to-Arabic translation.
Details on the datasets, prompts, and evaluation are presented in Appendix~\ref{appendixb}.

\subsection{Evaluating the Human Ability to Detect Machine-Generated Text}
\label{humanevaluation}
The gold standard for evaluating a model's language generation capability is human evaluation.
We presented 74 Arabic-speaking subjects from various social media with a survey designed to test the average-human ability to distinguish between machine-generated and human-written text and thus testing the model's ability to deceive a human subject.
The survey had a total of 8 news articles, 4 machine-generated using \textsc{AraGPT2}-Mega and 4 written by humans.
Each category was split into long and short text, which allows us to test the long-term generation coherency. 
In addition, the human evaluators are allowed to add justification for each answer.

\begin{figure}[!ht]
\centering
  \includegraphics[width=\columnwidth]{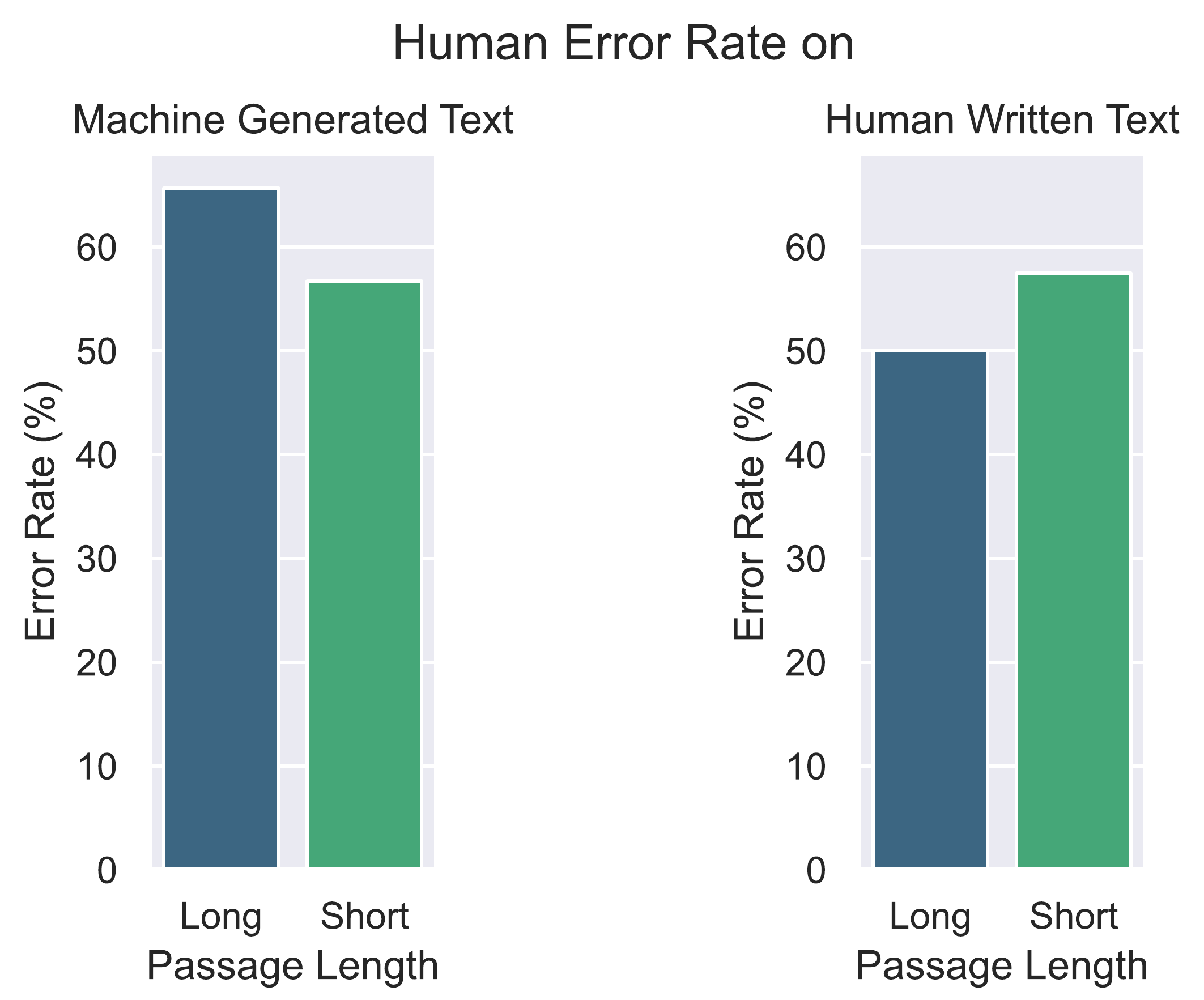}
  \caption{Survey results showing human error rates on machine generated \textit{(left)} and human written text \textit{(right)}\label{survey_res}}
\end{figure}

The survey results, Figure~\ref{survey_res}, show that \textsc{AraGPT2}-Mega successfully fooled approx. 60\% of the respondents, with longer passages having a higher error rate than short passages.
In the provided explanations, some subjects relied on punctuation mistakes, coherence, and repetition issues, while others spotted factual inaccuracies.
However, the results also show that humans were misclassifying human-written 50\% the time (chance level performance), while also citing factual inconsistencies, grammatical errors, and unusual writing styles\footnote{Survey results are available on our GitHub repository.}.

These surprising results show that \textsc{AraGPT2} can accurately generate human-like text while maintaining grammatical correctness that can fool the average reader.
It should be noted that there exist some tools, i.e. the Giant Language model Test Room (GLTR)~\cite{gehrmann2019gltr}, that allows humans to study the statistical distributional differences in text generated by GPT2-based models and human-written text.
Figure~\ref{gltr} in Appendix~\ref{appendixc} displays a visualization of token-level information created by GLTR with text generated by \textsc{AraGPT2} and on human-written articles.

\section{Automatic Detection of Machine Generated Text}
\label{sec:disc}
Large language models could have a significant societal impact if used for malicious purposes, such as automating the generation of misleading news articles, fake reviews, or high-quality phishing messages.
The survey in Section~\ref{humanevaluation}, showcases the failure of the average-human to consistently detect machine-generated text, which motivates the problem of automatic detection of \textsc{AraGPT2}-generated text.
Related work on the detection of machine-generated text by~\citet{jawahar-etal-2020-automatic} indicates that automatic detectors like the GROVER-detector~\cite{zellers2019defending} and the RoBERTA-detector~\cite{solaiman2019release} have better success than human evaluators.
In addition, previous work on detecting Arabic GPT2~\cite{khooligpt2} auto-generated tweets, achieved 98.7\% accuracy, by fine-tuning an \textsc{AraBERT}v0.1~\cite{antoun2020arabert} based classifier~\cite{harrag-etal-2020-bert}.

Our detector is based on the pre-trained \textsc{AraELECTRA}~\cite{antoun2020araelectra} model, which we fine-tuned on a dataset created by combining 1500 human-written news articles, with 1500 articles generated by \textsc{AraGPT2}-Mega.
For article generation, we only provided the model with a short prompt of 25 words.
We created two versions of the dataset, one with short texts (150 tokens) and one with long texts (500 tokens), in order to evaluate the impact of the text's length.

Fine-tuned \textsc{AraELECTRA} achieves 98.7\% and 94.9\% F1-score on long and short text respectively\footnote{The trained model will be publicly available in our repository}, which indicates that longer text is easier to detect than short text.
The high scores achieved by \textsc{AraELECTRA} can be explained by the fact that machine-generated text tends to be more predictable compared to human-written text (see Appendix~\ref{appendixc}, Fig.~\ref{gltr}).
The difference in text predictability can be easily exploited by a language model to detect machine-generated text.
Another contributing factor is that \textsc{AraELECTRA} was pre-trained on the exact same dataset as \textsc{AraGPT2}.

\section{Conclusion}
\label{sec:conc}
\textsc{AraGPT2} is the first advanced Arabic language generation model based on the transformer architecture. 
The model was trained on the largest publicly available collection of filtered Arabic corpora.
The model was evaluated using the perplexity measure which measures how well a probability model predicts a sample.
Results show that \textsc{AraGPT2} is able to produce high quality Arabic text that is coherent, grammatically correct and syntactically sound.

It is important to note that \textsc{AraGPT2}, like many ML models, has ethical implications and can be used maliciously i.e. automatic fake news generation, modeling the dataset inherent biases...
To help detect misuse of the model, a detector model that is tasked to detect output generated by \textsc{AraGPT2} is also released.
More importantly, our hopes that publicly releasing \textsc{AraGPT2} will open up doors for new research possibilities for the Arabic NLP community.

\section*{Acknowledgments}
This research was supported by the University Research Board (URB) at the American University of Beirut (AUB), and by the TFRC program for providing free access to cloud TPUs.
Many thanks to As-Safir newspaper for the data access, and also thanks to Nick Doiron for the insightful discussions.

\bibliography{references.bib}
\bibliographystyle{acl_natbib}

\onecolumn
\appendix
\section{Generated Samples from \textsc{AraGPT2}}
\label{appendixa}
\begin{figure*}[!h]
  \includegraphics[width=\textwidth]{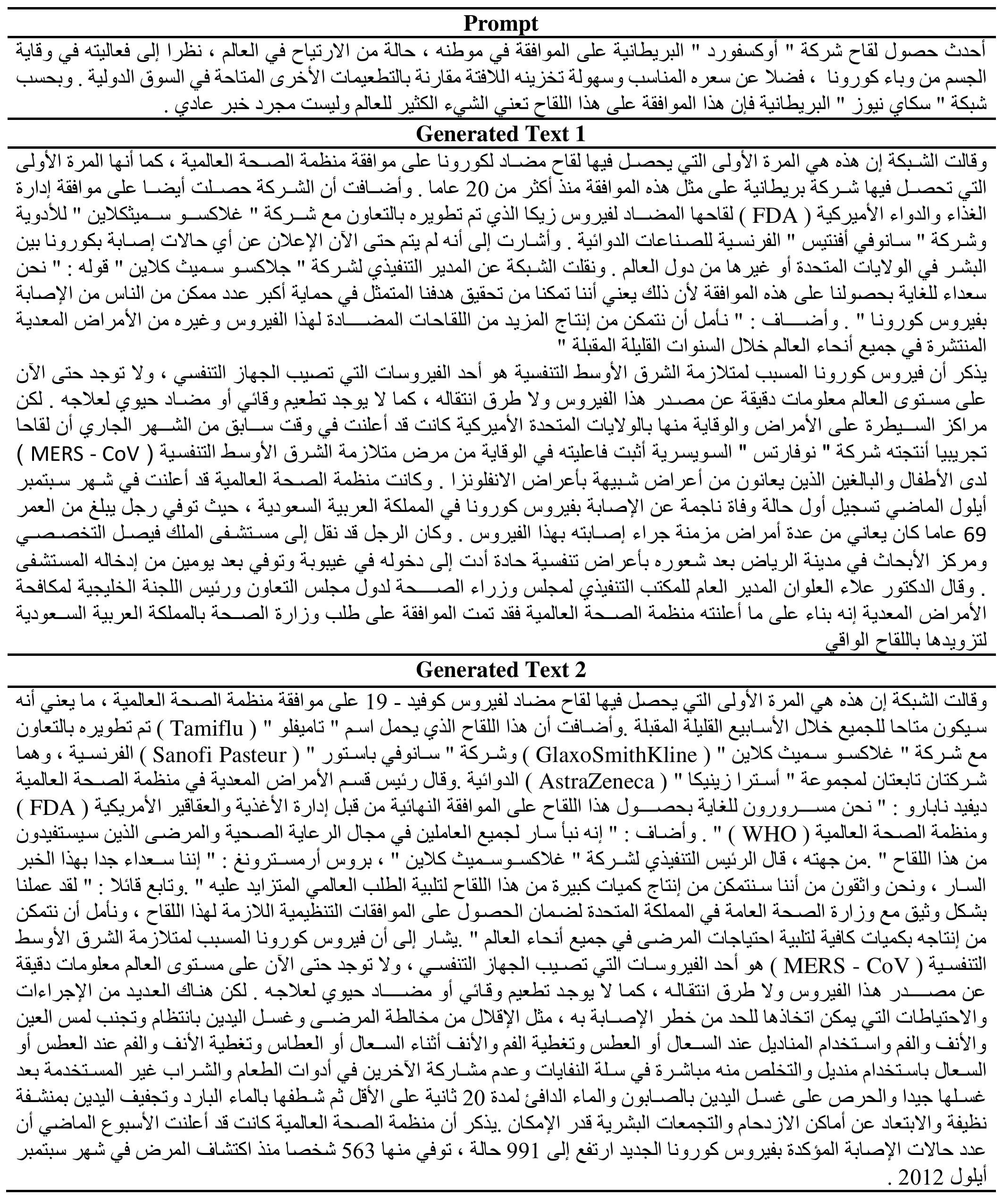}
  \caption{Random unseen context about coronavirus vaccine(top). Followed by two generated samples bu \textsc{AraGPT2}-mega. Generated text 1 ($top_p=0.95$), Generated text 2 ($top_p=1$)\label{prompt1}}
\end{figure*}
\newpage
\begin{figure*}[!h]
  \includegraphics[width=\textwidth]{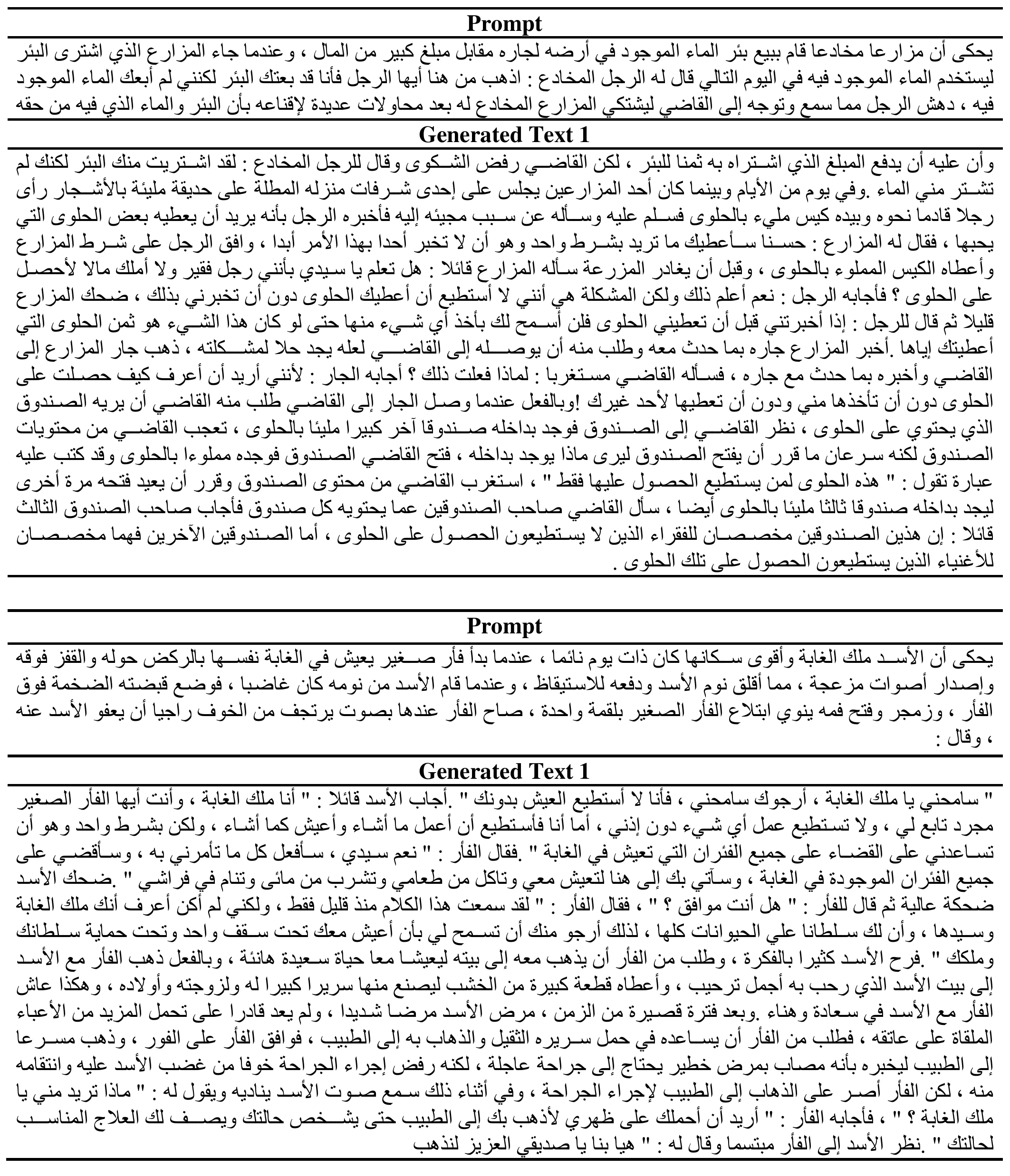}
  \caption{Random unseen contexts about children stories. Followed by a generated sample by \textsc{AraGPT2}-mega with $top_p=0.95$\label{prompt2}}
\end{figure*}
\newpage
\begin{figure*}[!h]
  \includegraphics[width=\textwidth]{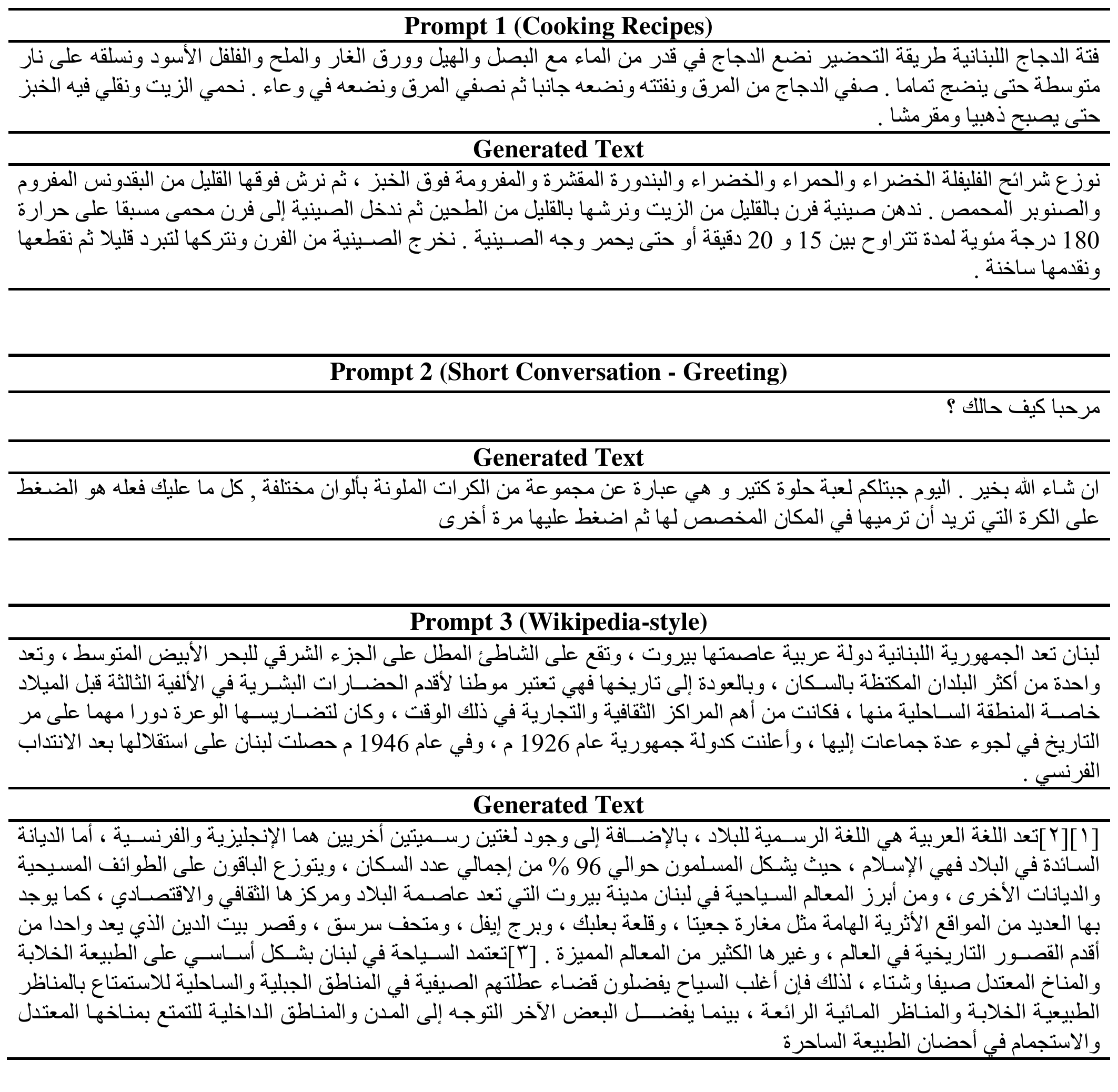}
  \caption{Random unseen contexts on miscellaneous topics. Followed by a generated sample by \textsc{AraGPT2}-mega with $top_p=0.95$\label{prompt3}}
\end{figure*}

\twocolumn
\newpage
\section{Zero-Shot Learning}
\label{appendixb}

\subsection{Question Answering}
In zero-shot factoid question answering, the information contained within the language model can be queried.
The model is tested on the Arabic examples from the TyDiQA~\cite{tydiqa} validation dataset (921 examples), and on the test set of ARCD~\cite{mozannar-etal-2019-neural} (702 examples).
Hence, the model os provided with the following prompt: \textit{``Answer the following question:}'' - ``\setcode{utf8}{\small{\<أجب عن السؤال التالي:>}''}
, followed by the question, then the phrase \textit{``The answer is}'' - ``\setcode{utf8}{\small{\<الجواب هو:>}''}.
It is also possible to append the phrase \textit{``in the year}'' - ``\setcode{utf8}{\small{\<في عام>}''}, if the expected answer is a year, as shown in Table~\ref{prompt}.

\begin{table}[!ht]
\centering
\caption{The input prompt for question answering\label{prompt}}
\resizebox{\columnwidth}{!}{%
    \begin{tabular}{r}
    \toprule
        \small{\<أجب عن السؤال التالي : متى عرضت أول حلقة من مسلسل>} \\
        \small{\<بافي قاتلة مصاصي الدماء ؟ الجواب هو في عام>}
        \\ \multicolumn{1}{l}{\textit{Answer the following question: When was the first }} 
        \\ \multicolumn{1}{l}{\textit{episode of the series Buffy the Vampire Slayer shown?}}
        \\ \multicolumn{1}{l}{\textit{The answer is in the year}} \\
    \bottomrule
    \end{tabular}
}
\end{table}

The answer length is set to be the same as the gold answer length, and a repetition penalty is applied as in CTRL~\cite{keskar2019ctrl}, which penalizes the probability scores of previously generated tokens.
A `no repeat tri-gram' strategy that inhibits the model from generating the same tri-gram more than once has also been employed.
Note that the context passage is not provided, which forces the model to rely only on the information gained during pretraining.

The model achieves a 3.93\% exact-match score and an F1-score of 14.51\% on TyDiQA, and 4.07\% exact-match score and 13.88\% F1-score on ARCD.
Since exact-match and F1-score misses answers that are correct but are worded differently (as shown in Table~\ref{correct}).
A subset of 500 answers from the best TyDiQA run is selected, and scored manually.
Manual scoring shows that \textsc{AraGPT2} correctly answered 24.6\% of the questions.
The model was particularly good in countries and capitals question, year of birth and death, and some geography.
Yet it was failing mostly on questions about quantities i.e. population counts, area, age...
The pre-defined answer length negatively affected the generated answers in some cases, which is a limitation of the current approach.

\begin{table}[!ht]
\centering
\caption{Examples of correct answers that have zero exact match score.\label{correct}}
\resizebox{\columnwidth}{!}{%
    \begin{tabular}{c|r}
        \toprule
        \textbf{Question} & \<من هو ألفرد نوبل ؟> \\ & \textit{Who is Alfred Nobel?} \\
        \midrule
        \textbf{Predicted Answer} & \RL{\< مخترع الديناميت ، ومخترع>} \\
        & \textit{Inventor of the dynamite, and the inventor of} \\
        \midrule
        \textbf{Ground Truth} & \RL{مهندس ومخترع وكيميائي سويدي} \\
        & \textit{An engineer and an inventor and a Swedish chemist}\\
        \bottomrule
        \bottomrule
        \textbf{Question} & \<متى تاسس الاتحاد الدولي لكرة القدم ؟> \\ & \textit{When was the FIFA founded?} \\
        \midrule
        \textbf{Predicted Answer} & \<1904 م .> \\
        & \textit{1904 AD} \\
        \midrule
        \textbf{Ground Truth} & \RL{21 مايو من العام 1904} \\
        & \textit{21 May of the year 1904}\\
        \bottomrule
        \bottomrule
        \textbf{Question} & \<من هو إدغار ديغا ؟> \\ & \textit{Who is Edgar Degas?} \\
        \midrule
        \textbf{Predicted Answer} & \<أنه فنان تشكيلي فرنسي ، ولد في باريس عام> \\
        & \textit{He is a French visual artist, born in } \\
        \midrule
        \textbf{Ground Truth} & \RL{فنان تشكيلي و رسام و نحات فرنسي} \\
        & \textit{Visual artist and painter and sculptor}\\
        \bottomrule
    \end{tabular}
}
\end{table}

\subsection{Translation}
A experiments has also been conducted to test the translation capability of \textsc{AraGPT2} by appending the prompt \textit{``What is the translation of this sentence ?:}'' - ``\setcode{utf8}{\small{\<ما هي ترجمة هذه الجملة ?:>}''}
to the sentence from the source language, in order to induce the translation behavior of the model.
We then apply greedy decoding to get the generated target sentence.
Evaluation is performed on 5000 randomly selected pairs from the English-Arabic Tatoeba~\cite{TIEDEMANN12.463} dataset.
The model achieved only 1.32 BLEU score\footnote{Using the sacrebleu scorer~\cite{post-2018-call}}. The low score is due to the scarce representation of English words in the vocabulary, since most words were split into single characters.
Additionally, given that the prompt design greatly affects the model's zero-shot performance, our prompt design might have been sub-optimal.
Nevertheless, this negative result encourages research into prompt engineering for Arabic language models, which we leave as future work.

\onecolumn
\newpage
\section{GLTR Analysis and Visualizations}
\label{appendixc}

\begin{figure}[!ht]
     \centering
     
     \begin{subfigure}[b]{\textwidth}
         \centering
         \includegraphics[width=\textwidth,trim={0 0 0 0.1cm},clip]{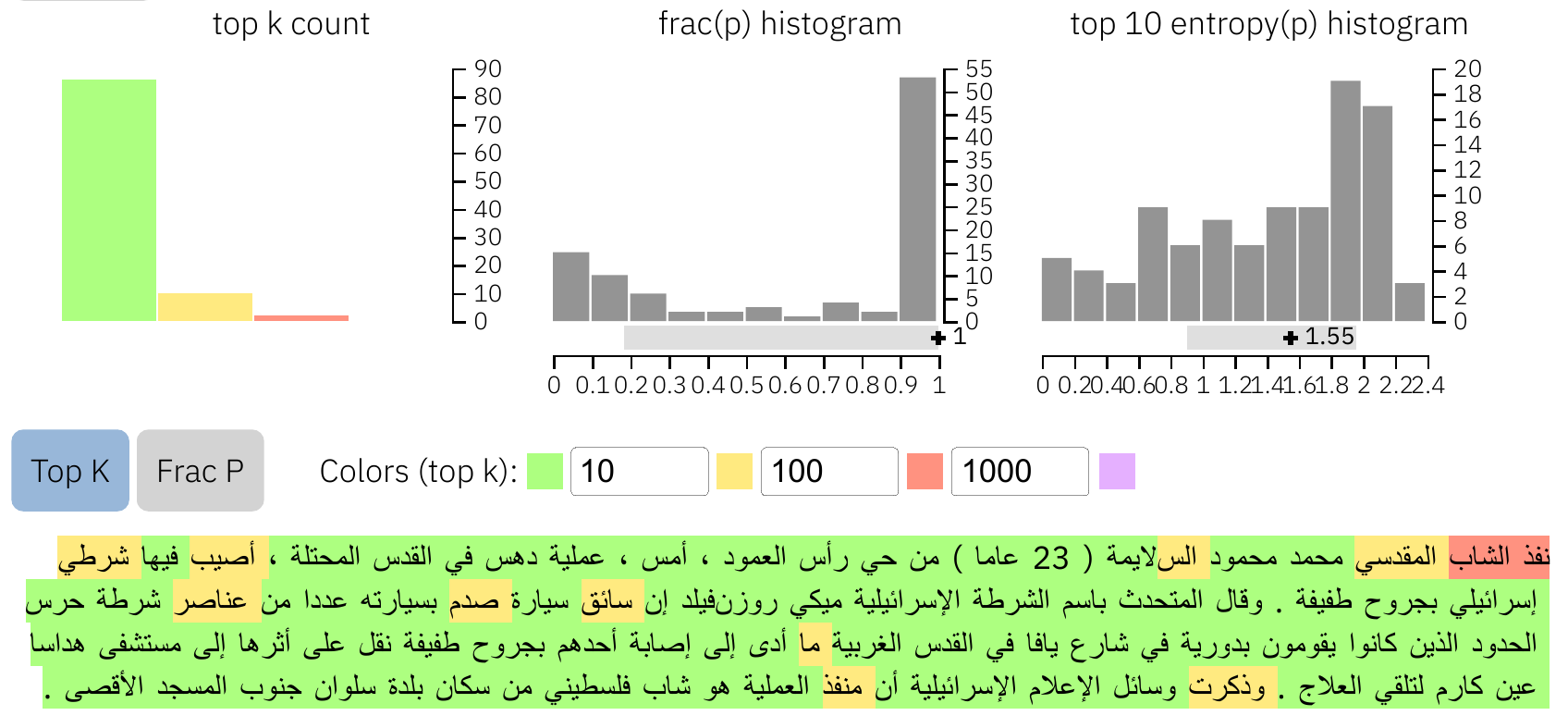}
         \caption{Text generated by \textsc{AraGPT2}-Mega. The first sentence is the human-written prompt}
         \label{fig:y equals x}
     \end{subfigure}
     
     \vspace{\floatsep}
     
     \begin{subfigure}[b]{\textwidth}
         \centering
         \includegraphics[width=\textwidth]{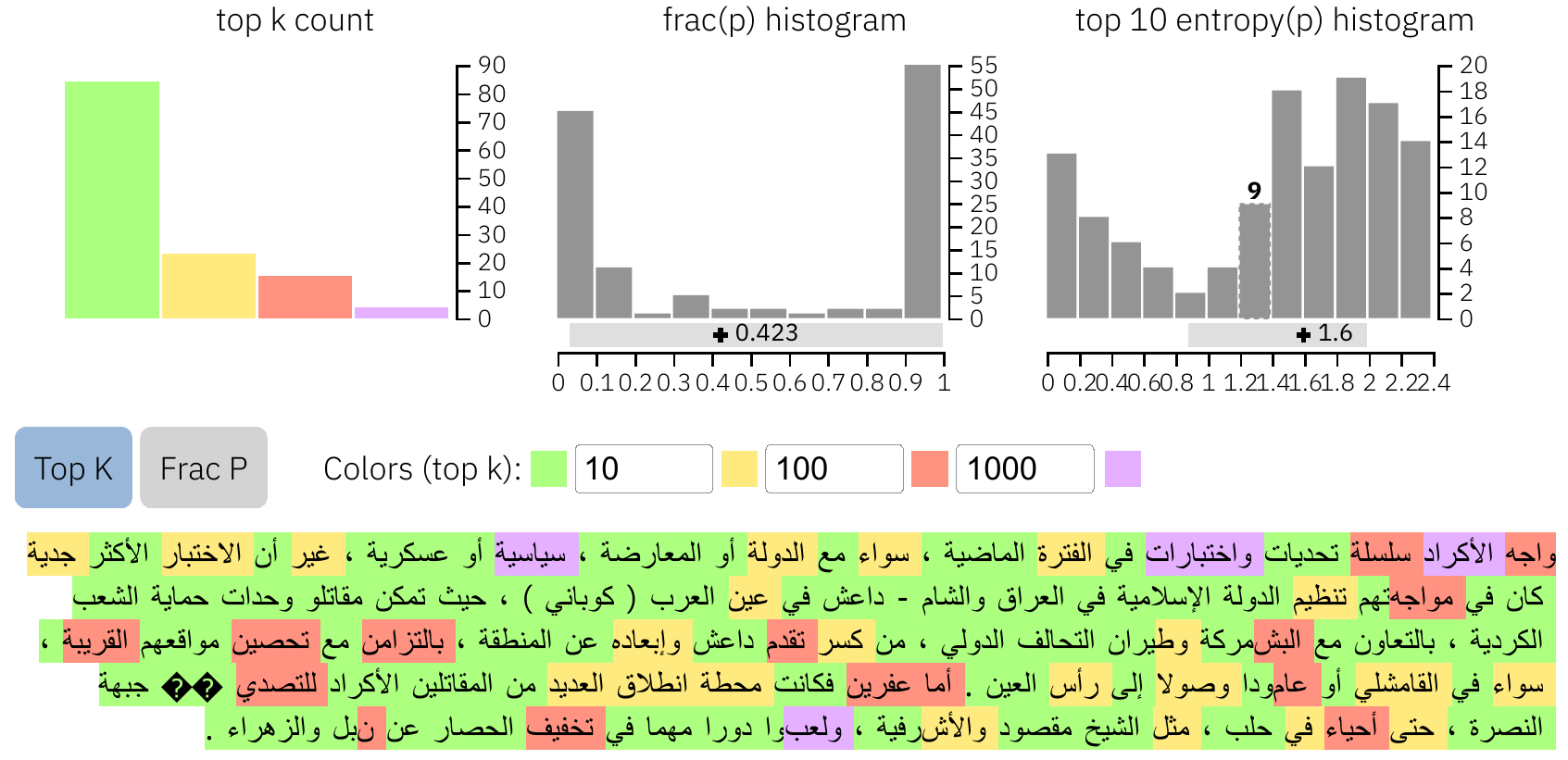}
         \caption{Human-Written Text}
         \label{fig:three sin x}
     \end{subfigure}
     \caption{It is clear that the machine generated text in (a) is mostly green and yellow highlighted, while in the human-written text, (b), an increase in red and purple highlighted words can be noticed. P.S.: We use \textsc{AraGPT2}-base as the backend model in GLTR}
     \label{gltr}
     
\end{figure}

\end{document}